\documentclass[sigconf, nonacm]{acmart}

\renewcommand{\tilde}[1]{\widetilde{#1}}
\renewcommand{\hat}[1]{\widehat{#1}}
\usepackage{mathtools,optidef}
\mathtoolsset{showonlyrefs}
\usepackage[ruled,vlined,linesnumbered]{algorithm2e}
\usepackage{subcaption}

\AtBeginDocument{%
  }

\copyrightyear{2026}
\acmYear{2026}
\setcopyright{cc}
\setcctype{by-nc-nd}
\acmConference[WWW '26]{Proceedings of the ACM Web Conference 2026}{April 13--17, 2026}{Dubai, United Arab Emirates}
\acmBooktitle{Proceedings of the ACM Web Conference 2026 (WWW '26), April 13--17, 2026, Dubai, United Arab Emirates}
\acmPrice{}
\acmDOI{10.1145/3774904.3792841}
\acmISBN{979-8-4007-2307-0/2026/04}

\begin{document}

\title{BanditLP: Large-Scale Stochastic Optimization for Personalized Recommendations}


\author{Phuc Nguyen}
\affiliation{%
  \institution{LinkedIn}
  \city{San Francisco}
  \country{USA}}
\email{honnguyen@linkedin.com}

\author{Benjamin Zelditch}
\affiliation{%
  \institution{LinkedIn}
  \city{New York}
  \country{USA}}
\email{bzelditch@linkedin.com}

\author{Joyce Chen}
\affiliation{%
  \institution{LinkedIn}
  \city{Sunnyvale}
  \country{USA}}
\email{joychen@linkedin.com}

\author{Rohit Patra}
\affiliation{%
  \institution{LinkedIn}
  \city{New York}
  \country{USA}}
\email{ropatra@linkedin.com}

\author{Changshuai Wei}
\authornote{Corresponding author}
\affiliation{%
  \institution{LinkedIn}
  \city{Seattle}
  \country{USA}}
\email{chawei@linkedin.com}

\renewcommand{\shortauthors}{Phuc et al.}

\begin{abstract}
We present BanditLP, a scalable multi-stakeholder contextual bandit framework that unifies neural Thompson Sampling (TS) for learning objective-specific outcomes with a large-scale linear program (LP) for constrained action selection at serving time. The methodology is application-agnostic, compatible with arbitrary neural architectures, and deployable at web scale—with an LP solver capable of handling billions of variables. Experiments on public benchmarks and synthetic data show consistent gains over strong baselines. We apply this approach in LinkedIn’s email marketing system and demonstrate business win, illustrating the value of integrated exploration and constrained optimization in production.
\end{abstract}




\maketitle
\sloppy 

\section{INTRODUCTION}

Most contextual-bandit (CB) based recommender systems (RS) focus on optimizing a single objective, typically the utility of end users, such as clicks or engagement metrics. However, in real-world platform ecosystems, multiple stakeholders are involved. For example, buyers and sellers on Amazon, artists and listeners on Spotify, viewers and content creators on YouTube or TikTok, and users, drivers, and restaurants on UberEats or DoorDash all form distinct stakeholder groups. Within an organization, there may also be multiple stakeholders. We see within LinkedIn email marketing system that the stakeholders include LinkedIn users, business lines within LinkedIn (e.g. B2C vs. B2B), and the platform itself may have different objectives and constraints. From the user-perspective, there are conflicting objectives: maximizing conversions while maintaining a low unsubscribe rate to preserve long-term reach. From the business lines perspective, we need fair representation—for example, to prevent the algorithm from over-promoting B2C products (which often generate quicker, higher-volume conversions) at the expense of B2B offerings. At the platform level, constraints such as maximum the number of emails sent to each member per week or the total volume of emails must be enforced. 

These systems require careful balancing utilities of all these stakeholders, and call for multi-stakeholder contextual bandits that balances exploration and exploitation while jointly managing potentially conflicting objectives across these groups. We can frame this multi-stakeholder problem as maximizing a primary cumulative reward subject to multi-level constraints that safeguard stakeholders’ interests in every round. This formulation is broadly applicable in real-world systems that involves batch recommendations, where operational constraints such as budget, capacity limits, and latency targets coexist with fairness, minimum exposure, or safety requirements. For instance, in e-commerce platforms like Amazon, recommendation policies must balance user engagement against seller exposure and inventory or delivery capacity limits. In content platforms such as TikTok or YouTube, models must allocate impressions fairly across creators while optimizing viewer satisfaction and respecting latency or throughput constraints~\cite{dong2024measuring, su2024long}. In LinkedIn’s email marketing system, recommendation policies must satisfy user-level frequency caps, business-line-level representation requirements, and platform-level performance guardrails and volume limits simultaneously. These settings all share a common structure: optimizing a primary objective while satisfying constraints across multiple stakeholders and system levels.

While prior research has advanced contextual bandits under constraints or multiple objectives, none provides a solution suited to large-scale applications with complex, multi-level stakeholder constraints. Existing approaches based on linear or multi-armed bandits~\cite{amani2019linear, khezeli2020safe, saxena2020thompson} cannot capture complex, high-dimensional relationships between users, items, and contexts. Methods developed for constrained or safe bandits typically handle only specific types of constraints or a single constraint level~\cite{daulton2019thompson, ge2025multi}, making them difficult to apply in multi-level constraints settings. On the other hand, large-scale recommender systems~\cite{surer2018multistakeholder, lu2025optimizing} that combine predictive models with LP solvers lack exploration, which amplifies selection bias and leads to suboptimal long-term outcomes. To our knowledge, no existing method jointly achieves i) the expressiveness to model complex, high-dimensional relationships, ii) the flexibility to satisfy interdependent multi-level stakeholder constraints, and iii) the scalability required for web-scale deployment.

Our work addresses this gap by proposing \emph{BanditLP}, an algorithm which combines neural Thompson sampling (TS) for exploration–exploitation and expressive modeling with a large-scale linear program (LP) to balance multi-stakeholder objectives under constraints. In our approach, rewards and costs are first estimated and sampled via neural TS; then, at each decision round, an LP sub-routine selects actions that satisfy multi-level constraints corresponding to different stakeholders.

In summary, the primary contributions of this paper are:
\begin{itemize}
    \item \textbf{Novel methodology}: An algorithm, BanditLP, that couples neural TS with large-scale LP for multi-stakeholder systems. The method is highly generalizable: it applies across applications, is compatible with arbitrary deep learning architectures for modeling objectives, supports a rich set of constraints, and scales to billions of recommendations. We validate BanditLP on public benchmarks and synthetic datasets, showing gains over established baselines.
    \item \textbf{Large-scale application with real-world impacts}: We deploy BanditLP in LinkedIn’s large-scale email marketing system. Online experiments show significant improvements across multiple business objectives, leading to production deployment. To our knowledge, this is the first large-scale email marketing RS that combines contextual bandits with constrained optimization and demonstrates business impact via online A/B testing.
    \item \textbf{Insights from deployment and A/B test}: We describe several modifications for large-scale deployment, including speeding up and scaling neural TS, tuning and monitoring the degree of exploration, calibrating Thompson-sampled probabilities, and configuring the LP. We also detail an online experiment design that avoids effect dilution and data leakage, enabling a fair comparison between recommenders with and without exploration.
\end{itemize}

\section{RELATED WORK}

Contextual bandit algorithms have emerged as a powerful framework for personalized recommendation systems, with applications spanning healthcare, finance, and information retrieval. This demonstrates the versatility of bandit-based approaches in solving real-world recommendation challenges~\cite{li2010contextual, qin2014contextual, zhang2020conversational, bouneffouf2020survey}. Recent advances~\cite{allesiardo2014neural, zhou2020neural, ban2021ee, zhu2023scalable} leverage the representational power of neural networks, with some algorithms achieving provable regret guarantees~\cite{zhou2020neural, ban2021ee} and others showing strong empirical success~\cite{zhu2023scalable}. However, traditional CB formulations optimize a single scalar reward, which limits their applicability in multi-objective or multi-stakeholder recommendation environments.

A rich body of work has studied constrained multi-armed bandits, where costs are introduced alongside rewards, and the learner aims to maximize cumulative reward while satisfying constraints. One line of research focuses on knapsack-style constraints, where interactions terminate once a constraint is violated~\cite{badanidiyuru2018bandits, liu2022non}, or where cumulative constraints must be respected across all rounds~\cite{guotriple}. These differ from our setting, where constraints should be satisfied in every round. Another branch closely related to our problem considers per-round constraints, often requiring the construction of safe decision sets~\cite{amani2019linear, khezeli2020safe}, which can become computationally expensive when constraints are complex. In ~\cite{saxena2020thompson}, a global constraint imposes a minimal level of performance in each round, and a linear bandit is combined with an LP sub-routine to select actions. Our work shares similarities with~\cite{saxena2020thompson} in combining a contextual bandit method with linear programming (LP) to select arms that balance constraints at each round. However, we extend this idea to a richer contextual bandit setting with more complex reward and cost functions and a larger set of constraints. Moreover, prior work in constrained multi-arm or linear bandits primarily focuses on regret analysis under simplified settings and on offline experiments. In contrast, we aim to show how contextual bandits can handle complex relationships between context and rewards, as well as large-scale operational constraints in web applications.

The literature on contextual bandits for multi-objective and multi-stakeholder settings is comparatively limited. Some approaches convert multiple objectives into a single scalar objective through scalarization~\cite{busa2017multi, tekin2018multi, mehrotra2020bandit}, but this is unsuitable when certain objectives (e.g., minimum selection or spend budgets) must be treated as hard constraints. Other methods only accommodate specific types of constraints. For instance,~\cite{daulton2019thompson} constrains total performance relative to a baseline level, while~\cite{ge2025multi} solves the combinatorial bandits for budget allocation in advertising using Bayesian hierarchical models. These formulations are not directly applicable to our problem, which involves potentially complex and multi-level constraints. In~\cite{harsha2025practical}, an algorithm combines general-purpose learning models with discrete constrained optimization to generate a pool of safe solutions and employs inverse gap weighting to sample actions. However, in large-scale ecosystems with billions of user–item combinations, maintaining such a pool becomes computationally prohibitive.

Our method is motivated by recent trends in recommender systems that increasingly integrate predictive models with scalable LP solvers to enforce multi-objective trade-offs and operational constraints at decision time. For example,~\cite{surer2018multistakeholder} uses base recommender scores as LP inputs to balance buyer- and provider-side objectives, and~\cite{lu2025optimizing} addresses platform-level business constraints. Other applications employ LP to satisfy a variety of constraints, including optimizing email volume~\cite{basu2020eclipse}, ensuring fairness in ranking~\cite{geyik2019fairness}, promoting diversity in network growth~\cite{ramanath2022efficient}, and managing marketplace budget allocations~\cite{ramanath2022efficient}. However, these approaches rely solely on supervised models trained for predictive accuracy on logged data, without any mechanism for exploration. This lack of exploration often reinforces feedback loops—models repeatedly recommend items similar to those already popular, leading to “rich get richer” dynamics—and ultimately results in suboptimal long-term performance~\cite{li2010contextual, su2024long, swaminathan2015batch}. Furthermore, unlike bandit-based recommenders, such systems are prone to the cold-start problem~\cite{wang2017biucb}. A key distinction of our approach is the use of neural Thompson sampling, which naturally balances exploration and exploitation to mitigate selection bias, handle uncertainty, and adapt to new information for improved long-term outcomes.

\section{METHODOLOGY}

\subsection{BanditLP}\label{sec:banditlp}

For ease of exposition, similar to \cite{surer2018multistakeholder}, we assume that there are three stakeholders in the multi-stakeholder ecosystem: the platform, the provider (i.e., cohorts of items, sellers, business lines) and the user, although our result is generalizable to more stakeholders. In each round, the RS selects a set of items for each user. Let $u=1,\cdots,U$ index users, $i=1,\cdots,I$ index items, and $L$ be a set of providers such that for each $i \in \mathbb{I}_l$ the item $i$ belongs to provider $l$. In each round $t = 1,...,T$, the RS observes a contextual feature $z^{\mathrm{user}}_{u,t}$ for a user $u$, and $z^{\mathrm{item}}_{i,t}$ for an item $i$. If the RS recommends item $i$ for user $u$, it receives a set of feedback drawn from an unknown distribution, where $r_{u,i,t} = r_t(z_{u,i,t})$ with $z_{u,i,t} = [\,z^{\mathrm{user}}_{u,t};\,z^{\mathrm{item}}_{i,t}\,]$ is the primary reward representing a true-north business metric, and $\{c^{(k)}_{u,i,t} = c^{(k)}_t(z_{u,i,t})\}_{k=1}^K$ are $K$ costs and secondary feedback relevant to all stakeholders in the ecosystem. 

We frame this multi-stakeholders problem as maximizing the cumulative primary reward as the main objective subject to constraints that safeguard stakeholders' interests in every round. We can characterize the constraints in a three-stakeholder ecosystem as:
\begin{itemize}
    \item Platform-level constraints: a constraint involves feedback for all users and items.
    \item Provider-level constraints: a constraint is calculated over items belonging to each provider. This can also generalize to include constraints calculated over users belonging to a cohort.
    \item User-level (or item-level) constraints: a constraint is for each user.
\end{itemize}
Some examples of platform-level constraint are total daily ads budget for an ads platform, maximum unsubscription rate for an email service, or minimum platform-wide revenue performance. This constraint is often found in safe bandit works \cite{saxena2020thompson, daulton2019thompson}. Some examples of provider-level constraint are minimum percentage of promoted products from each seller in an online marketplace, or maximum deviation from a provider's budget, which are often found as fairness constraints in fair bandit works~\cite{sinha2024assessing}. User-level constraint can be the maximum number of recommendations a user may get~\cite{wei2024neural}, or item-level constraint may be minimum number of selection or a budget~\cite{li2019combinatorial}.


The goal is to design a policy that:
\begin{maxi!}|l|[2]
  {x_{u,i,t}}{%
    \sum_{t=1}^T\sum_{u=1}^U\sum_{i=1}^I x_{u,i,t}\, r_{u,i,t} \label{eqn:obj}%
  }{}{} 
  \addConstraint{\sum_{u}\sum_{i} x_{u,i,t}\, c^{(1)}_{u,i,t}}
               {\le C_1 \label{eqn:cost-platform}}{\forall\,t}
  \addConstraint{\sum_{u}\sum_{i\in \mathbb{I}_l} x_{u,i,t}\, c^{(2)}_{u,i,t}}
               {\le C_{2l} \label{eqn:cost-provider}}{\forall\,l,\,t}
  \addConstraint{\sum_{i \in I} x_{u,i,t}\, c^{(3)}_{u,i,t}}
               {\le C_3 \label{eqn:cost-user}}{\forall\,u,\,t}
  \addConstraint{0 \le x_{u,i,t} \le 1}{}{\forall\,u,\,i,\,t}
\end{maxi!}

\noindent where $x_{u,i,t}$ denotes the probability of selecting item $i$ for user $u$ at time $t$. 

We observe that a satisfactory policy can be obtained by solving a subroutine linear program (LP) in every round. Specifically, in the LP formulation, in the current round $t$, the goal is to maximize a primary business objective $\sum_{u}\sum_{i} x_{u,i,t}\, r_t(z_{u,t},i)$, subject to platform-level, provider-level, and user-level constraints as included in \eqref{eqn:cost-platform}, \eqref{eqn:cost-provider}, \eqref{eqn:cost-user} respectively. As long as the number of providers in $L$ is small, the problem is computationally feasible. 

In practice, since the true $r_{u,i,t},\, c^{(k)}_{u,i,t}$ are unknown, we assume that they can be estimated via Bayesian neural networks (see Section~\ref{sec:neuralee}) $\tilde{r}_t(z_{u,i,t})$ for the primary reward and $\tilde{c}^{(k)}_t(z_{u,i,t})$ for the costs and secondary reward. Exploration can be achieved through Thompson sampling given neural posterior predictive distributions. A policy satisfying the objective and constraints above can be computed based on solving the LP with the estimated rewards and costs ~\cite{basu2020eclipse}. After serving those recommendations, we update the estimators $\tilde{r}_t(z),\, \tilde{c}^{(k)}_t(z)$ to obtain improved utility value estimates before the next round. We summarize BanditLP in Algorithm \ref{alg:banditlp}.

\begin{algorithm}[t]
\caption{BanditLP}\label{alg:banditlp}
\DontPrintSemicolon
\SetKwInOut{Input}{Input}
\SetKwInOut{Initialize}{Initialize}

\Input{Time horizon $T$; providers $l=1,\dots,L$ with item index sets $\{\mathbb{I}_l\}_{l=1}^L$ partitioning $\{1,\dots,I\}$; three-stakeholders constraint thresholds $C_1$, $\{C_{2l}\}_{l\in L}$, $C_3$.}
\Initialize{Initialize Bayesian neural networks $\tilde{r}_0(z)$ and $\{\tilde{c}^{(k)}_0(z)\}_{k=1}^3$.}

\For{round $t=1$ \KwTo $T$}{
    \For{each user $u=1,\dots,U$ and item $i=1,\dots,I$}{
        Observe contexts $z_{u,i,t}$. Draw Thompson samples from neural posterior predictive distributions based on Laplace approximation:
        $$\hat r_{u,i,t} \sim \mathbb{P}_r\!\big(\cdot \,\big|\, \tilde r_{t-1}(z_{u,i,t}), \mathcal D_{1:t-1}\big)$$
        $$\hat c^{(k)}_{u,i,t} \sim \mathbb{P}_k\!\big(\cdot \,\big|\, \tilde c^{(k)}_{t-1}(z_{u,i,t}), \mathcal D_{1:t-1}\big),\,\forall k$$
    }
    \textbf{Solve the LP:}
        \begin{maxi!}|l|[1]
        {x_{u,i,t}}{%
          \sum_{u}\sum_{i} x_{u,i,t}\,\hat{r}_{u,i,t}
        }{}{}
        \addConstraint{\sum_{u}\sum_{i} x_{u,i,t}\,\hat{c}^{(1)}_{u,i,t}}{\le C_1}{}
        \addConstraint{\sum_{u}\sum_{i\in \mathbb{I}_l} x_{u,i,t}\,\hat{c}^{(2)}_{u,i,t}}{\le C_{2l}}{\quad\forall\,l}
        \addConstraint{\sum_{i} x_{u,i,t}\,\hat{c}^{(3)}_{u,i,t}}{\,\le C_3}{\quad\forall\,u}
        \addConstraint{0 \le x_{u,i,t} }{\le 1}{\quad\forall\,u,\, i.}
        \end{maxi!}
    \For{each user $u=1,\dots,U$}{
        Choose a set of items $A_{u,t}$, where an item $i$ is chosen based on $x_{u,i,t}$.\;
        Observe feedback
    $\,r_{u,i,t},\, \{c^{(k)}_{u,i,t}\}_{k=1}^3$ for $i\in A_{u,t}$.\;
    }
    Update the Bayesian neural networks $\tilde{r}_t(z)$ and $\{\tilde{c}^{(k)}_t(z)\}_{k=1}^3$ using the new observations.\;
}
\end{algorithm}

The proposed BanditLP method offers both versatility and scalability, making it suitable for a wide range of real-world applications. The multi-level constraint formulation generalizes numerous constrained bandit settings, encompassing item-level and group-level fairness, campaign-level budget control, and global minimum safety requirements, as well as arbitrary combinations of these. The neural Thompson sampling component can be paired with any neural network architecture, while the LP layer is fully decoupled from the prediction model, enabling flexible customization for diverse objectives and constraints. From a scalability perspective, the LP solver can efficiently handle billions of decision variables, and the neural network component can model complex, high-dimensional relationships. This modular architecture also supports incremental testing and staged deployment in production environments, facilitating reliable integration into large-scale systems.

\subsection{Neural Thompson Sampling}\label{sec:neuralee}

In this section, we briefly summarize our large-scale implementation of Bayesian neural networks to estimate unknown rewards and costs functions $r_t(z),\, \{c^{(k)}_t(z)\}_{k=1}^K$. We quantify the uncertainty in our estimators by approximating their intractable posteriors by multivariate Gaussian via the Laplace approximation. Laplace approximations are widely used for uncertainty quantification in recommendation system ~\cite{su2024long,joachims2018deep, daxberger2021laplace,foong2019between,raha2024computationally}.

Let us assume that for an input $z$, we have a neural network $f_\theta(z)$ to predict the probability of a binary reward $r$. Then under linearized Laplace approximation (LLA) of~\cite{daxberger2021laplace,foong2019between,raha2024computationally}, the posterior predictive distribution for a new observation $(z^{*},r^{*})$ in the binary classification setup will be $P(r^*=1|\mathbf{z}^*, \mathcal{D}_{train}) = \sigma(\mu(z^{*})),$ with
\begin{align} \label{Gen_Gauss:Newton:precision}
\mu(z^{*}) &\sim N\left(f_{\hat{\theta}_{MAP}}(z^{*}), \tau V \right)\\
V&:= \sigma_{0}^{2}+g(z^{*})^{T} \Omega^{-1} g(z^{*})\\
\Omega&:=\sum_{n=1}^{N}\,g(\mathbf{z}_{n}) \left( \nabla_{f}^{2}\, log \,p(\mathbf{r}_{n}|f)\Big|_{f=f_{\hat{\theta}_{MAP}}(\mathbf{z}_{n})}\right) g(\mathbf{z}_{n})^{T}
\end{align}
where $\tau$ is a temperature to scale the posterior covariance, $g(\cdot) := \nabla_\theta f_\theta (z^*)|_{\theta = \hat{\theta}_{MAP}}$,  $\nabla_{f}^{2}\, log \,p(r_{n}|f):=e^{f}/(1+e^{f})^{2}$, $\sigma(\cdot)$ is the sigmoid function, $\mathcal{D}_{\text{train}}$ denotes the training data set, and $\sigma^2_0$ is the variance on the Gaussian prior on $\theta$, and $\hat{\theta}_{MAP}$ are network weights learned from standard training with gradient descent.\footnote{For numeric reward or costs, we use Gaussian for posterior predictive distribution and use the Gauss-Newton matrix for Hessian for regression task, see~\cite{foong2019between} for more details.} 

In our production scale settings, $\Omega$ is often too large to invert directly, the literature focuses on lower dimensional approximation such as top $k$ eigen-pairs of $\Omega$\cite{nilsen2022epistemic}, a diagonal matrix ~\citep{zhang2020neural}, {sub-network approximation} that focuses on identifying a small subset $S$ of parameters~\cite{daxberger2021laplace} and use the sub matrix $\Omega_{S,S}$ to approximate $\Omega$, use the \textit{last few layers} of the network as the choice of $S$~\cite{riquelme2018deep,snoek2015scalable}. Our implementation adopts the last strategy.

For applications where a baseline supervised neural network model exists, it may be desirable to be able to quickly deploy or rollback on explore-exploit capability. Neural TS provides this advantage because we can apply LLA on a wide range of pre-trained neural network~\cite{daxberger2021laplace}. We discuss practical tips to select the number of the last layers of the network, as well as tune and monitor degree of exploration for web-scale applications in Section~\ref{sec:email-application-overlap-k}.

\subsection{Large-scale Linear Program solver}\label{sec:dualip}

To solve an LP with up to billions of variables, we turn to \textit{DuaLip}~\cite{basu2020eclipse,ramanath2022efficient,dualip}, a large-scale LP solver developed here at LinkedIn. We briefly outline the algorithm implemented in Dualip here. To begin, for a given round $t$, observe that the LP in Section~\ref{sec:banditlp} in can be written in the following form:
\begin{align}
    \mathop{\text{min}}_{x} \: r^\top x \quad s.t. \quad Dx \leq b, \quad x \in [0,1]^{UI} \quad \forall u, i, \label{eqn:dualip-lp}
\end{align}
where $x =(x_{1,1,t},\cdots,x_{U,I,t}) \in \mathbb{R}^{|U||I|}$ is the vector of decision variables, $r = -1 \left( r_t(z_{1,t}, 1)),\cdot, r_t(z_{U,t},I) \right) \in \mathbb{R}^{UI}$ is the coefficient vector for the decision variables in the objective, $D$ is the design matrix in the constraints, $b = (C1, C_{21},\cdots, C_{2L}, C3)^\top$ is the constraint budget vector.

Instead of solving the LP in~\eqref{eqn:dualip-lp} directly, we solve a perturbed version of the LP, which is a quadratic program (QP):
\begin{align}
    \mathop{\text{min}}_{x} \: r^\top x + \frac{\gamma}{2} x^\top x \quad s.t. \quad Dx \leq b, \quad x_{u,i,t} \in [0, 1] \quad  \forall u, i, \label{eqn:dualip-qp}
\end{align}
where $\gamma > 0$ is a regularization hyperparameter. Dualizing just the polytopal constraint $Dx \leq b$ leads to the partial Lagrangian dual
\begin{align*}
    g_\gamma (\lambda) &= \min_{x} \: \left\{ r^\top x + \frac{\gamma}{2} x^\top x + \lambda^\top (Dx - b) \right\},
\end{align*}
The problem is transformed to calculate $\text{max}_{\lambda}g_{\gamma}(\lambda)$ such that $\lambda \geq 0$, assuming strong duality holds. We can then solve the dual problem efficiently using first-order methods, by using parallelizable projections. For sufficiently small $\gamma$, this yields the same solution as the original LP.  More details on DuaLip can be found in~\cite{basu2020eclipse,ramanath2022efficient, dualip} and tips for implementation can be found in~\cite{wei2024neural}.

\subsection{Application in Email marketing at LinkedIn}\label{sec:email-application}
\subsubsection{Formulating the BanditLP problem}
We now turn to a large-scale application of BanditLP at LinkedIn: email marketing recommendation. In email marketing at LinkedIn, we use batch inference to make personalized, per‑member send decisions at massive scale across many campaigns, with decisions governed by eligibility and frequency controls. The primary business objective is to maximize total expected revenue subject to a global constraint on total marketing unsubscriptions, as well as minimum business-line-level selection and maximum member-level volume constraints each week. 

We apply the BanditLP formulation in Section~\ref{sec:banditlp} to this application. Let $u=1,\cdots,U$ and $i=1,\cdots,I$ index users and campaigns respectively. In any given week $t$, conditional on context $z_{u,i,t}$, we estimate the primary reward for sending campaign $i$ to member $u$ as $y^{\mathrm{conv}}_{u,i,t} \cdot y^{\mathrm{LTV}}_{u,i,t}$, where $y^{\mathrm{conv}}_{u,i,t}= f^{\mathrm{conv}}(z_{u,i,t})$ is an estimate for a short-term reward measured by conversions, $y^{\mathrm{LTV}}_{u,i,t}= f^{\mathrm{LTV}}(z_{u,i,t})$ is an estimate for a long-term reward measured by life-time value (LTV) associated with a conversion. We also estimate unsubscription as $y^{\mathrm{unsub}}_{u,i,t}= f^{\mathrm{unsub}}(z_{u,i,t})$. Here $f^{\mathrm{conv}}$ and $f^{\mathrm{unsub}}$ are all estimated with neural TS using LLA as described in Section~\ref{sec:neuralee}, coupled with probability calibration. The $f^{\mathrm{LTV}}$ model is maintained by a separate system and fitted by a supervised learning method. 

\begin{figure}[H]
    \centering
    \includegraphics[width=0.9\linewidth]{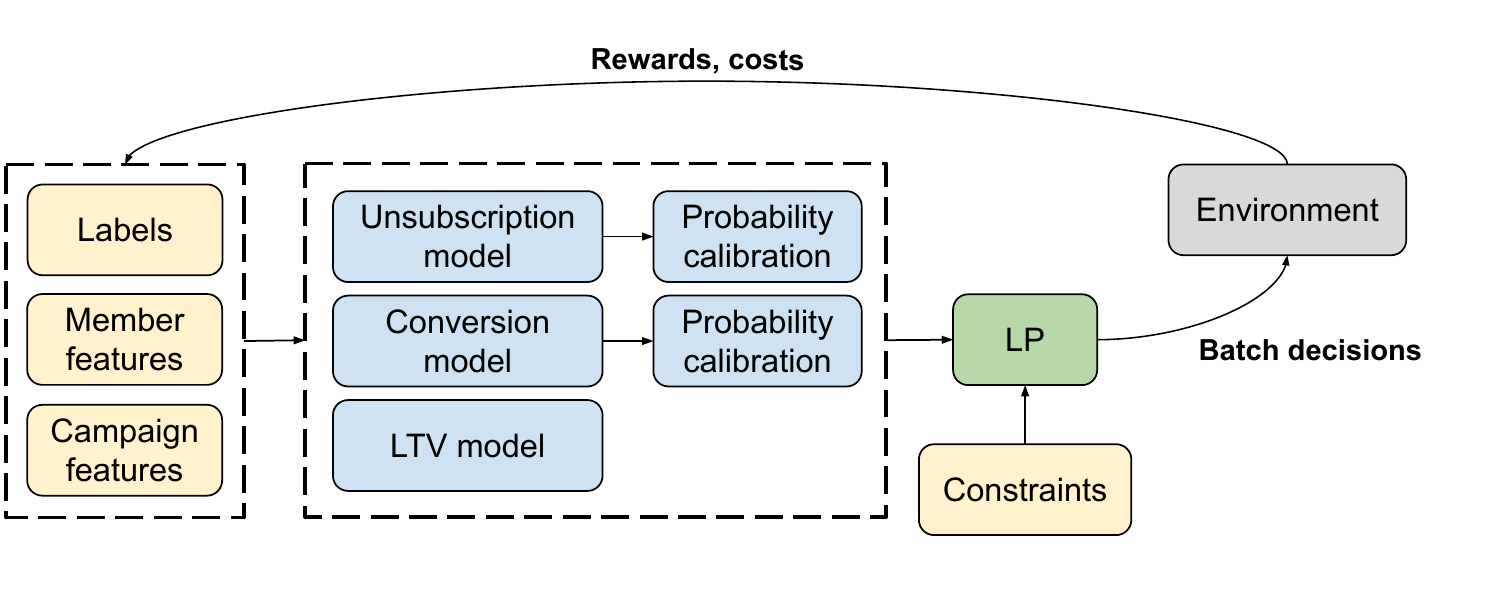}
    \caption{Overview of email marketing recommender system based on BanditLP at LinkedIn.}
    \label{fig:email-engine}
\end{figure}

Additionally, we have several constraints on the costs and send decisions that must be satisfied weekly: i) the total unsubscription should not be more than $C_{\mathrm{unsub}}$, ii) the minimum sends for campaigns for B2B products must be $C_{\mathbb{2B}}$, and the minimum sends for campaigns for B2C products must be $C_{\mathrm{2C}}$, and iii) the maximum number of campaigns that can be sent to a user is $C_{\mathrm{fcap}}$. The resulting BanditLP problem in each week is:
\begin{maxi!}|l|[2]
  {x_{u,i,t}}{
    \sum_{u}\sum_{i}
    x_{u,i,t}\,y^{\mathrm{conv}}_{u,i,t}\,y^{\mathrm{LTV}}_{u,i,t}
  }{\label{eqn:email-engine-obj}}{}
  \addConstraint{\sum_{u}\sum_{i}
    x_{u,i,t}\,y^{\mathrm{unsub}}_{u,i,t}}{\le C_{\mathrm{unsub}} \label{eqn:email-engine-unsub}}
  \addConstraint{\sum_{u}\sum_{i \in \mathbb{I}_{\mathrm{2B}}} - x_{u,i,t}}{\le -C_{\mathrm{2B}}}{}
  \addConstraint{\sum_{u}\sum_{i \in \mathbb{I}_{\mathrm{2C}}} - x_{u,i,t}}{\le -C_{\mathrm{2C}}}{}
  \addConstraint{\sum_{i} x_{u,i,t}}{\le C_{\mathrm{fcap}}}{\forall\,u}
  \addConstraint{0 \le x_{u,i,t} \le 1}{}{\forall\,u,\,i}
\end{maxi!}
Notice that the action probability variable $x$ has dimension $UI$, the product of the number of members and the number of campaigns, which can result in tens of billions of variables. We use \textit{DuaLip} to solve for $x_{u,i,t}$, the probability of sending campaign $i$ to user $u$ in week $t$. The main parameter to tune is $\gamma$, which controls the trade-off between speed and precision of the solver. Prior works~\cite{basu2020eclipse, wei2024neural} have discussed practical tips for implementing and tuning $\gamma$ that also work well in our application. Figure~\ref{fig:email-engine} shows the overview of the email marketing recommender based on BanditLP.

\subsubsection{Tuning and monitoring the degree of exploration}\label{sec:email-application-overlap-k}
We describe practical tips for tuning neural TS when its outputs feed an LP. First, the temperature parameter $\tau$ controls the exploration–exploitation trade-off. Because TS samples pass through an LP with complex constraints, doubling $\tau$ does not necessarily double the effective exploration downstream. We therefore propose an \emph{overlap-at-$K$} metric to measure the realized degree of exploration:
\begin{align}
    \mathrm{Overlap\text{-}at\text{-}}K \;=\; \frac{1}{N}\sum_{n=1}^{N}
    \frac{\bigl|\,S_n^{\mathrm{TS}}(K)\cap S_n^{\mathrm{exploit}}(K)\,\bigr|}{K},
\end{align}
where $N$ is the number of observations, and $S_n^{\mathrm{TS}}(K)$ and $S_n^{\mathrm{exploit}}(K)$ denote the top-$K$ actions for observation $n$ under the neural-TS policy and the same neural model without exploration, respectively. The metric quantifies how often the $K$ actions chosen by the TS policy agree with those from the exploit-only policy; values closer to $1$ indicate less exploration.

For our application, we set $K = C_{\mathrm{fcap}}$ and choose a target $p_{\mathrm{safe}} \in (0,1)$. We then grid-search over $\tau$ to achieve overlap-at-$K \ge p_{\mathrm{safe}}$ prior to deployment. Second, during model update, overlap-at-$K$ also serves to monitor distribution shift in model scores after refreshes, providing a safety check on exploration. Finally, we select the largest suffix of layers used for posterior-uncertainty approximation that still satisfies training and serving latency requirements.

\subsubsection{Calibrating neural Thompson sampled probabilities}
\begin{figure}[!htb]
    \centering
    \includegraphics[width=\linewidth]{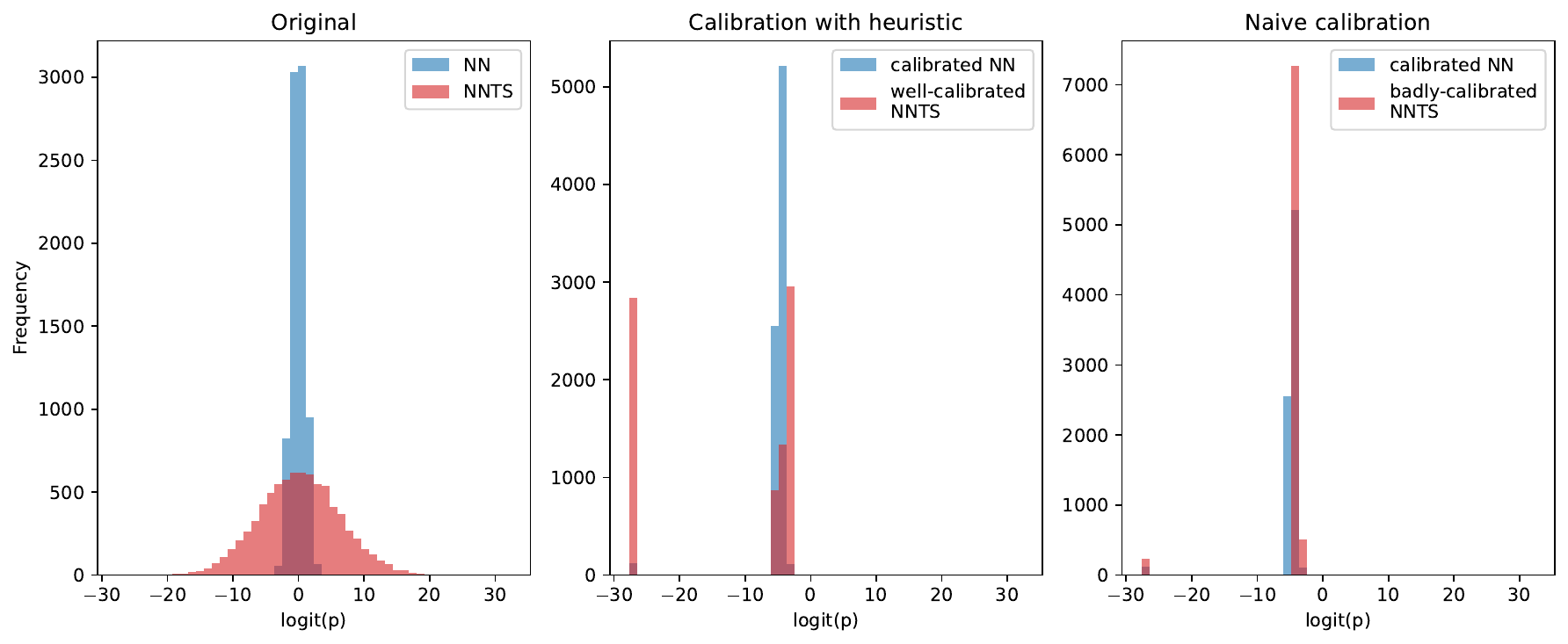}
    \caption{Calibration heuristic preserves exploration.}
    \label{fig:calibration}
\end{figure}
In our application, conversions and unsubscriptions are extremely rare, and negative labels are down-sampled for training. Thus, probability calibration is critical when predicted probabilities are used as inputs to an LP: the solver interprets these probabilities as quantitative values that directly constrain and guide optimization. We use an off-the-shelf isotonic regression, a non-parametric calibration technique that fits a monotone stepwise function mapping raw model scores to calibrated probabilities. 

Because uncertainty from the neural network cannot be directly propagated through isotonic regression, we adopt a practical heuristic. First, we fit a calibration model $g$ using predicted probabilities without exploration $\hat{p}$ and observed binary rewards $r$:
\[
g \;=\; \arg\min_{g \in \mathcal{M}} \sum_{i=1}^N \bigl(r_i - g(\hat{p}_i)\bigr)^2,
\]
where $\mathcal{M}$ denotes the space of monotone stepwise functions. At serving time, we obtain calibrated TS probabilities by applying $g$ to the Thompson-sampled predictions: $\tilde{p}^{\,\mathrm{TS}} = g\!\left(\hat{p}^{\,\mathrm{TS}}\right)$, where $\hat{p}^{\,\mathrm{TS}}$ are the model’s TS probabilities prior to calibration.

If we instead fit isotonic regression directly on TS probabilities, we observe a counterintuitive effect: increasing the temperature $\tau$ yields a \emph{more} concentrated distribution of calibrated probabilities. This arises because isotonic regression shrinks noisy TS draws toward a monotone mean function, effectively dampening exploration. Figure~\ref{fig:calibration} illustrates this anti-exploration behavior for naive TS calibration and the improvement under the proposed heuristic.

\section{EXPERIMENTS}


\subsection{Offline experiment} \label{sec:offline-exp}
\subsubsection{Baseline methods}
We compare BanditLP against standard contextual bandit methods and multi-objective recommender system without exploration. To our knowledge, there is no other work the combines a contextual bandit with an LP sub-routine for multi-stakeholders setting:
\begin{itemize}
    \item NNTS: a contextual bandit with strong predictive power using neural Thompson sampling but does not enforce constraints. 
    \item LinUCB-LP: an extension of LinUCB, a classical contextual bandit, by enforcing constraints during action selection by solving the LP in \eqref{eqn:dualip-lp}, thereby satisfying constraints in each round. However, LinUCB is limited in modeling complex (nonlinear) reward and cost functions.
    \item NN-LP: a supervised baseline that uses neural networks for deterministic predictions and solves the LP in \eqref{eqn:dualip-lp} during action selection (no exploration).
\end{itemize}


\subsubsection{Datasets}
We evaluate BanditLP on both a public benchmark and a synthetic dataset. The synthetic dataset allows us to select the number of constraints affecting the LP, the number of providers, and other characteristics of the recommendation problem. We generate random user and item features, assign items to mutually exclusive sets (providers), and define reward and costs as functions of user and item features. Appendix~\ref{apx:synthetic-setup} details the data-generation process. 

For the public benchmark, we use the Open Bandit Dataset (OBD)~\cite{saito2020open}, a real-world logged bandit dataset. We utilize the random-policy subset, where 34 products were randomly recommended to $>400$K users with binary click feedback. While OBD natively supports off-policy evaluation via the replay method, to enable efficient optimization of the LP sub-routine we construct an imputed reward matrix. Specifically, we randomly sample 10K users and fill unobserved user–item rewards using $K$-Nearest Neighbors ($K{=}10$): for each unseen item, we identify the 10 most similar users who interacted with it, compute their average CTR, and draw a Bernoulli sample accordingly. The resulting dense reward matrix serves as a controlled proxy for simulation studies and makes the LP tractable. We use the imputed clicks as the reward. We define synthetic costs similarly to the synthetic dataset. We also assign products evenly to disjoint sets (providers).

\subsubsection{Evaluation setup}
All models are initialized using training data collected by a \emph{biased logging policy} that selects items from a restricted region of the feature space. This reflects selection bias common in real-world RS (e.g., “rich-get-richer”) and cold-start effects for new items. Details for the synthetic setup appear in Appendix~\ref{apx:synthetic-setup}. For the OBD benchmark, the biased logging policy only recommends the top $30\%$ items uniformly at random.

To set constraints, we run a random policy and record its global and per-set costs, then set constraint targets as multipliers of these baselines. Intuitively, global costs may be targeted below a random policy (e.g., $70\%$), while per-set costs (e.g., fairness exposure) may be bounded relative to a perfectly balanced random policy (e.g., within $20\%$). For $T$ rounds, each model interacts with the environment by selecting actions for a set of users and items, observing rewards and costs, and updating via incremental training on its own feedback. We repeat the simulation for $S$ runs and report mean global reward, global-constraint violation, and per-set constraint violation with $95\%$ confidence intervals over all rounds. For the synthetic dataset, we set $T{=}30, \, S{=}50$, and for the OBD benchmark, we set $T{=}50, \, S{=}20$.

\subsubsection{Results}

\begin{figure}[!htb]
  \centering
  \includegraphics[width=.9\linewidth]{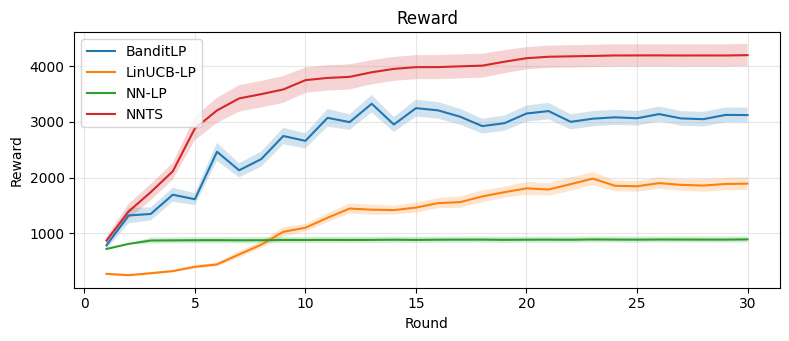}\par\vspace{0.6em}
  \includegraphics[width=.9\linewidth]{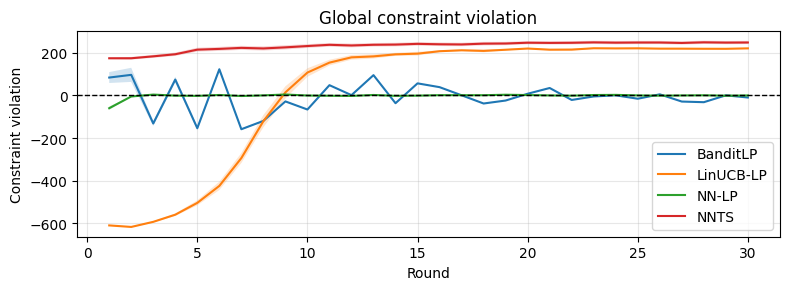}
  \includegraphics[width=.9\linewidth]{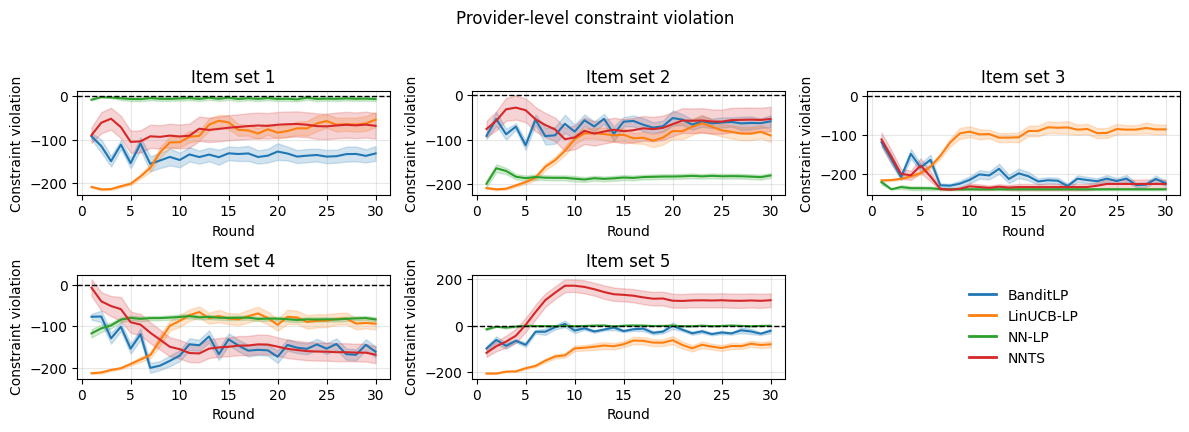}
  \caption{Synthetic experiments. \textbf{Top:} cumulative reward over rounds. \textbf{Middle:} global constraint violation (no violation if $\le 0$). \textbf{Bottom:} provider-level constraint violations across item sets (no violation if $\le 0$). Shaded regions show 95\% CIs over 50 runs.}
  \label{fig:synthetic-results}
\end{figure}

\begin{figure}[!htb]
  \centering
  \includegraphics[width=.9\linewidth]{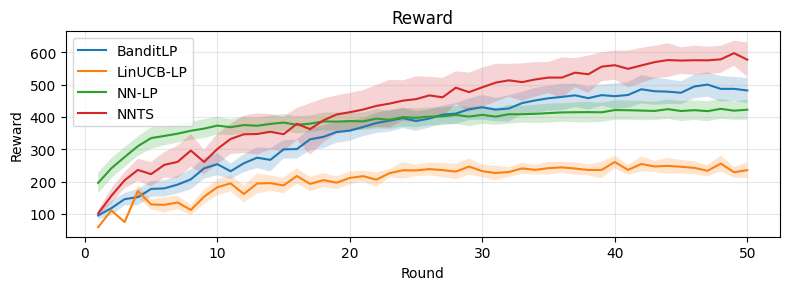}
  \includegraphics[width=.9\linewidth]{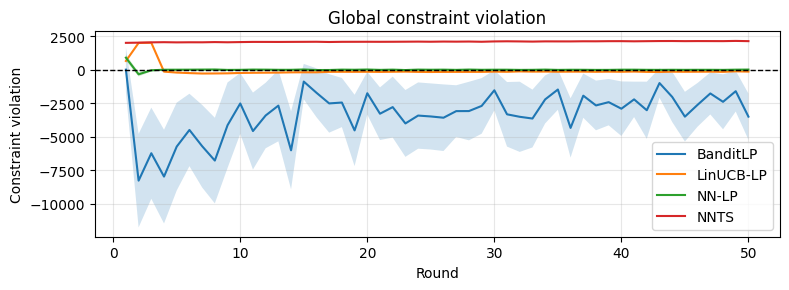}
  \includegraphics[width=.9\linewidth]{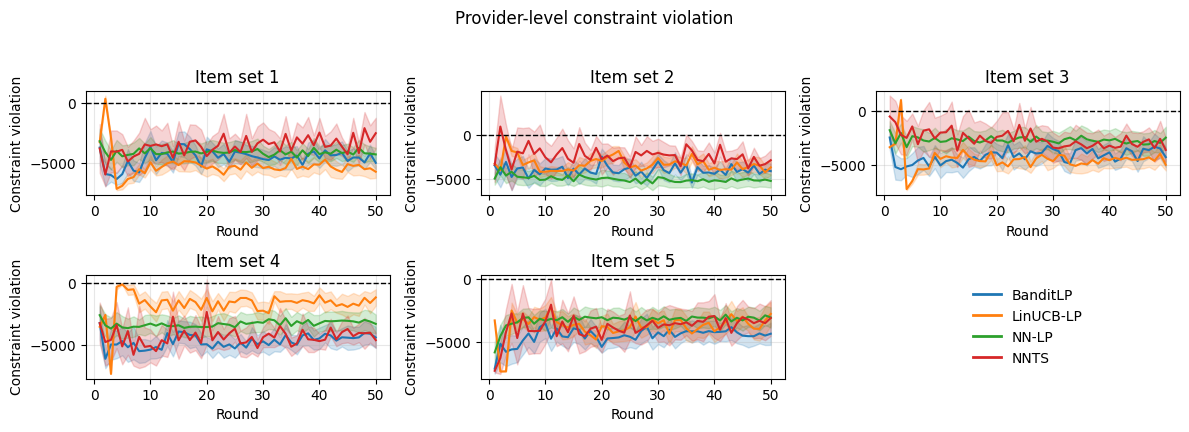}
  \caption{Open Bandit Dataset (imputed matrix) experiments. \textbf{Top:} cumulative reward over rounds. \textbf{Middle:} global constraint violation (no violation if $\le 0$). \textbf{Bottom:} provider-level constraint violations across item sets (no violation if $\le 0$). Shaded regions show 95\% CIs over 20 runs.}
  \label{fig:imputed-results}
\end{figure}

BanditLP method combines predictive accuracy with exploration and constraint satisfaction. Across both datasets, it achieves the best \emph{reward–feasibility} trade-off, maintaining violations near or below zero while delivering high reward (see Figure~\ref{fig:synthetic-results} and Figure~\ref{fig:imputed-results}). When the logged data to initialize the reward and costs models have selection bias, NNLP policy plateaus early as deterministic exploitation cannot escape selection-bias traps, yielding lower reward in the long-run. As expected, NNTS policy, which does not have an LP sub-routine, attains the highest reward but consistently violates global and per-set constraints. LinUCB-LP policy underperforms in reward while at the same time shows global constraint violation. Because the LP is solved using estimated rewards and costs, LinUCB's limited expressiveness to model non-linear cost functions leads to constraint misspecification and violations. 

\subsection{Online experiment}

\subsubsection{Design} We design an online A/B test to compare the performance of the BanditLP email marketing RS described in Section~\ref{sec:email-application} against a fully exploitative baseline. The control is a RS that only utilizes the same supervised learning models for $f^{\mathrm{conv}}, f^{\mathrm{unsub}}, f^{\mathrm{LTV}}$ but without Thompson sampling, and an LP to optimize for business constraints. The experiment therefore isolates BanditLP’s ability to \emph{induce exploration in the LP solutions} by injecting uncertainty into the estimated rewards and costs that enter the LP.

Because the email marketing system only covers a small percentage of all LinkedIn users, we carefully define the experiment population to reduce effect dilution. Rather than include all members, we restrict the population to the union of users who are eligible for weekly email recommendations during the test period. Concretely, we first assign all users to treatment or control, with assignments fixed for the duration of the experiment. Each week, when the set of eligible users is produced, those users are routed to their preassigned policy. For analysis, we include only users who were actually served by either the control or BanditLP RS during the experiment window.

\begin{figure}[h]
  \centering
  \includegraphics[width=0.8\linewidth]{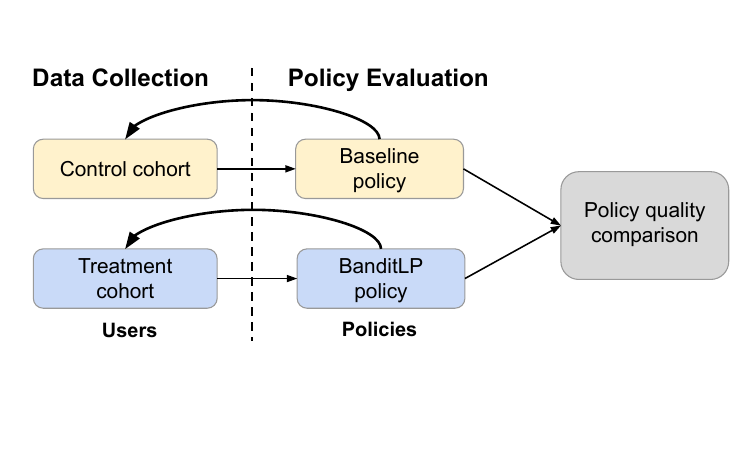}
  \vspace{-.4in}
  \caption{Data-diverted experiment setup to measure benefits of exploration.}
  \label{fig:online-exp-setup}
\end{figure}

Additionally, to avoid data leakage when A/B-testing a RS without exploration against one with exploration, we use a data-diverted experiment setup similar to \cite{su2024long}. Specifically, users random treatment assignments are enforced in both training and serving. Before launching the experiment, we train both RSs only on pre-experiment logs. During the experiment period, each RS may update using only data collected from its own cohort. This ensures there is no cross-learning and leaking the benefit of exploration from the treatment to the control policy. Figure~\ref{fig:online-exp-setup} illustrates the data-diverted experiment setup. We run the experiment for seven weeks, during which both the control and BanditLP RSs update on a weekly cadence with new interactions and feedback from the environment. This ensures the experiment can measure the effect of exploration on model learning.

We track revenue as the long-term, true north business metric, conversion rate as a short-term signal for reward, and unsubscription rate as our guardrail metric. In terms of statistical analysis, it is worth noting that the long-term revenue metric has zero inflation and heavy tail~\cite{wei2024neural}. Thus, we use the two-sample $t$-tests (Welch’s) as well as consider nonparametric zero-truncated tests~\cite{wei2025beyond} to robustly detect changes in this metric. 

\subsubsection{Results}

\begin{table}[h]
  \caption{Result of Online A/B Test}
  \label{tab:ab-result}
  \begin{tabular}{cl}
    \toprule
    Metric & Relative improvement ($95\%$ CI)\\
    \midrule
    Revenue & $+3.08\%  \quad (+0.17\%,\, +5.99\%)$\\
    Conversion rate & $-0.01\% \quad (-0.26\%,\, +0.23\%)$\\
    Unsubscription rate & $-1.51\% \quad (-2.25\%, -0.72\%)$\\
    \bottomrule
  \end{tabular}
\end{table}

The A/B test results using two-sample $t$-tests (Table~\ref{tab:ab-result}) show BanditLP produces statistically significant positive lift in long-term revenue metric ($+3.08\%$), and statistically significant reduction in unsubscription rate ($-1.51\%$). The short-term reward metric, conversion rate, shows a very small and not statistically significant change ($-0.01\%$). This is interesting because it suggests a trade-off between short-term and long-term gains: exploration may not immediately increase conversions, but it leads to higher sustained revenue over time. Additionally, the results show, by adding exploration to the estimates of rewards and costs using neural TS, BanditLP can effectively induce exploration in the LP solutions

\subsection{Ablation analysis}

Beyond demonstrating BanditLP's effectiveness in production via an online A/B test, we conduct ablation studies to measure its sensitivity to the quality of the rewards and costs estimators and levels of exploration. We seek to address the broad question: How will the LP behave given different estimators quality? In these analyses, we use our email marketing formulation for illustration.

We simulate BanditLP in a single round across varying levels of exploration $\tau$  and prediction model quality $q$. We generate realistic ground truth data and then model the data using varying degrees of model quality and exploration levels. For each pair $(q, \tau)$ of model quality and exploration level, we solve the LP, obtaining an allocation decision and associated downstream quantities. Precise details of the simulation setup are provided in Appendix~\ref{app:spa}. 

\noindent We seek to answer the following research questions:
\begin{enumerate} 
\item[\textbf{RQ1:}] What risks are introduced to constraints satisfaction due to exploration and model quality?
\item[\textbf{RQ2:}] How does exploration affect the stability of the LP's allocation decision?
\end{enumerate}

\begin{figure}[H]
  \centering
  \includegraphics[width=.8\linewidth]{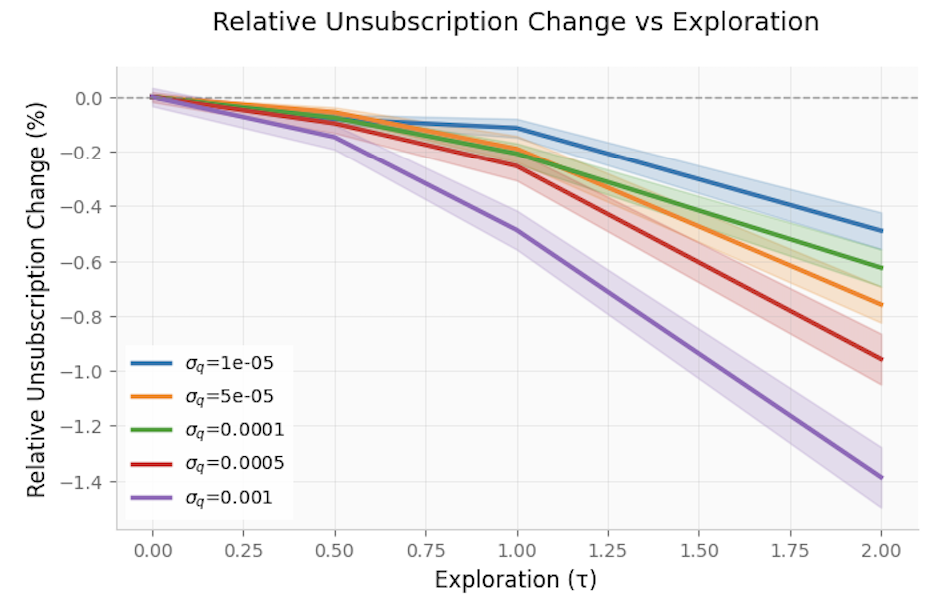}
  \vspace{-.1in}
  \caption{Relative unsubscription change under increasing degrees of exploration.}
  \label{fig:ablation-unsub}
\end{figure}

\textbf{RQ1 (Constraint Satisfaction).} Figure~\ref{fig:ablation-unsub} shows relative unsubscription change $\textrm{RUC}(q, \tau) = \frac{\bar{\mathrm{U}}(q, \tau) -\bar{\mathrm{U}}(q, 0)}{\bar{\mathrm{U}}(q, 0)} \cdot 100\%$. Unsubscriptions remain stable (within ±2\% of baseline) across all configurations, indicating the constraint remains effectively binding despite exploration noise. 

\begin{figure}[H]
  \centering
  \includegraphics[width=.8\linewidth]{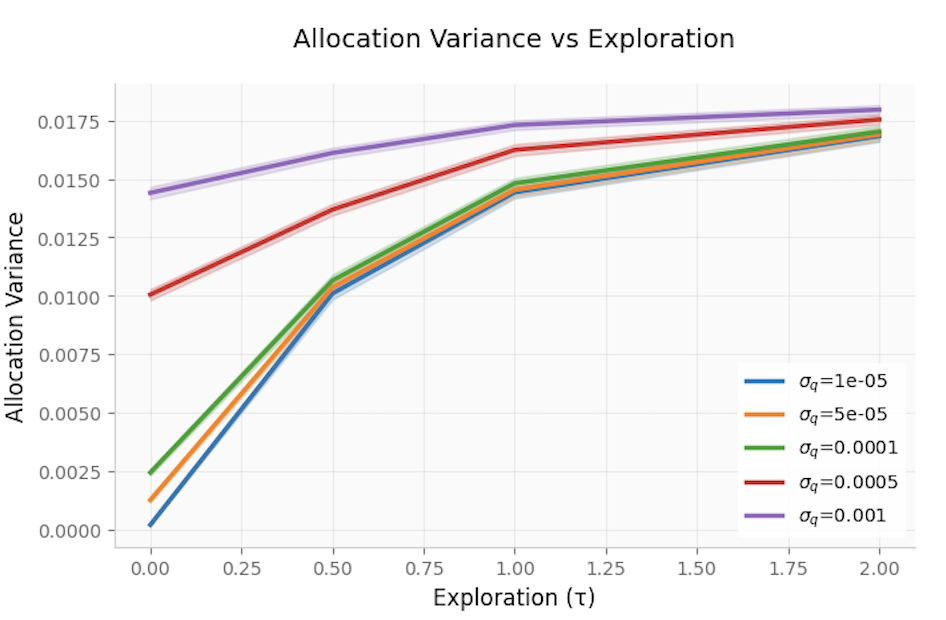}
  \vspace{-.1in}
  \caption{Total allocation variance change under increasing degrees of exploration.}
  \label{fig:ablation-allocation}
\end{figure}

\textbf{RQ2 (Allocation Stability).} Figure~\ref{fig:ablation-allocation} shows allocation variance $\textrm{AV}(q, \tau) = \frac{1}{|\mathcal{U}| |\mathcal{I}|}\sum_{u,i} \textrm{Var}\left(X_{u,i}\right)$. Variance increases with exploration, with high-quality models showing greater sensitivity. This occurs because low-quality models already contain substantial baseline instability, making exploration's marginal contribution relatively small.

\section{CONCLUSION AND FUTURE WORK}

This work introduces a general algorithm for multi-stakeholder contextual bandits that combines neural Thompson Sampling with a large-scale linear program to maximizing a primary reward subject to constraints that safeguard stakeholders’ interests in every round. All objectives are first estimated and sampled via neural TS, and the final recommendations are chosen by a decoupled LP that enforces stakeholders' constraints at scale. The system is application-agnostic, compatible with arbitrary deep learning architecture to estimate rewards and costs, supports a rich set of constraints, and scales to billions of recommendations. We validate the approach on public benchmarks and synthetic data against established baselines. In production, we demonstrate BanditLP's strengths in an application to LinkedIn email marketing system, balancing users, business lines, and platform goals under real constraints. The application delivers business impact through online A/B test. We describe deployment lessons such as probability calibration, tuning and scaling neural TS, improving LP convergence, and setting up bandit experiment to avoid data leakage.

The current formulation assumes no interaction effects across marketing units, and performance relies on accurately calibrated predictions feeding the LP. Moreover, end-to-end behavior depends on solver throughput and feasibility under tight constraints. Future work includes real-time inference and online learning to close the loop continuously in production, further automation of calibration and diagnostics for stability at scale, and extending the method to broader recommender settings and constraint families while preserving the separation between learning and optimization.

\begin{acks}
We thank Licurgo Benemann De Almeida, Kenneth Tay, John Bencina, Shruti Sharma, Rajni Sharma, Sumukh Sagar Manjunath, Curtis Seaton, Shawn Bianchi, and Alyssa Falk for their support and collaboration on the email application and A/B test.
\end{acks}

\bibliographystyle{ACM-Reference-Format}
\bibliography{refs}

\appendix

\section{Offline experiment}
\subsection{Synthetic experiment setup} \label{apx:synthetic-setup}
Here we describe in detail the synthetic data generation setup for offline experiments in Section~\ref{sec:offline-exp}. We set number of user each round to $U=500$, and number of items (fixed) to $I=100$, with items assigned to $L=5$ disjoint sets of items. To generate a multi-stakeholder contextual bandit problem, at each round, we sample user features $z_u \sim N(0, I), \, z_u \in \mathbb{R}^{d_u}$ and item features $z_i \sim N(0, I), \, z_i \in \mathbb{R}^{d_i}$, with $d_u = d_i = 10$. The true dominant reward and cost functions are:
\begin{align}
r_{u,i} &\sim N(\mu_r(z_u^\top \beta_u + z_i^\top \beta_i),\, \sigma^2_r), \\
c^{(1)}_{u,i} &\sim N(\mu_{c}(z_u^\top \beta^{(1)}_u + z_i^\top \beta^{(1)}_i),\, \sigma^2_{c}),\\
c^{(2)}_{u,i} &\sim N(\mu_{c}(z_u^\top \beta^{(2)}_u + z_i^\top \beta^{(2)}_i),\, \sigma^2_{c})
\end{align}
where the reward is centered around a sum of sigmoids of linear transformations of the features $\mu_r(v) := \sigma(-4v + 5) + 5\sigma(5v + 5) + 0.1 \sin(2v),$, and the costs are centered around $\mu_{c}(v):= 1 + 0.1 \tanh(v/2)$. We set $\sigma^2_r = \sigma^2_c = 0.1$, and sample linear weights $\beta_u, \beta_i, \beta^{(1)}_u, \beta^{(2)}_u, \beta^{(1)}_i, \beta^{(2)}_i \sim N(0, 0.6^2)$. 

We implement the biased logging initial policy as selecting actions with $z_u^\top \beta_u + z_i^\top \beta_i < 0$ uniformly at random. We set user-level constraint to be $2$. We simulate the global and per-set constraints by first running a random policy that selects up to two item for 5000 users, producing a set $A_{random}$. We then calculate the global constraint as $C_1 = 0.8 \sum_{u=1}^{5000}\sum_i c^{(1)}_{u,i}1_{u,i \in A_{random}}$ and per-set constraints as $C_{2l} = 1.5 \sum_{u=1}^{5000}\sum_i c^{(2)}_u 1_{u,i \in A_{random}}$.

\section{Ablation and sensitivity analysis} 

\label{app:spa}

Here we describe in detail the ablation simulation and analysis. We first simulate how different levels of exploration affect LP allocation decisions in a single period, across varying levels of prediction model quality $q$. We generate synthetic data under the following mechanism:
\begin{align}
\textrm{pConv}_{u,i} &\sim \textrm{Beta}(2, 18) \cdot 10^{-2},\\
\textrm{pUnsub}_{u,i} &\sim \textrm{Beta}(1, 9) \cdot 10^{-1},\\
\textrm{LTV}_{u,i} &\sim \textrm{Gamma}(2, 150).
\end{align}
We think of these as the true, unknown values that we estimate via prediction models.

Second, we generate model predictions for these ground truth values, which are noisy estimates of them. We introduce four model qualities $q$, each of which has an associated noise parameter $\sigma_q$ (higher quality models have lower noise parameter values). The model predictions are given by: 
$$\operatorname{logit}(\widehat{\textrm{pConv}}_{u,i}^q) = \operatorname{logit}(\textrm{pConv}_{u,i}) +\epsilon_q, \quad\epsilon_q \sim  \mathcal{N}(0,\sigma_q^2)$$

And similarly for $\widehat{\textrm{pUnsub}}_{u,i}^q$. 

Third, given the model predictions, we compute the prediction uncertainties: $U_{{\textrm{Conv}}_{u,i}}^q = \tau \gamma|\widehat{\textrm{pConv}}_{u,i}^q- \textrm{pConv}_{u,i}| + \gamma_0$. In practice we would need to estimate these uncertainties as well. Given them, we obtain the explored/perturbed predictions:

$$\widetilde{\textrm{pConv}}_{u,i} \sim \textrm{Beta}\left(\alpha_{\textrm{Conv}}^{q}, \beta_{\textrm{Conv}}^{q} \right),$$

where $\alpha_{\textrm{Conv}}^{q} = \widehat{\textrm{pConv}} \cdot n_{\textrm{Conv}}$, $\beta_{\textrm{Conv}}^{q} = (1-\widehat{\textrm{pConv}}) \cdot n_{\textrm{Conv}}$, and $n_{\textrm{Conv}} = \frac{\widehat{\textrm{pConv}} (1-\widehat{\textrm{pConv}} )}{({U_{{\textrm{Conv}}_{u,i}}^{q}})^2} - 1$. These choices ensure $\mathbb{E}\left[ \widetilde{\textrm{pConv}}_{u,i} \right] = \widehat{\textrm{pConv}}$ and $\mathrm{Var}\left(\widetilde{\textrm{pConv}}_{u,i} \right) \approx ({U_{{\textrm{Conv}}_{u,i}}^{q}})^2$. Here $\tau$ is a parameter which controls the degree of exploration (higher $\tau$ leads to more exploration). And similarly for unsubscription probabilities.

The LP we solve is:

\begin{maxi!}|l|[2]
  {x_{u,i}}{
    \sum_{u\in\mathcal U}\sum_{i\in\mathcal I}
    x_{u,i} \widetilde{\mathrm{pConv}}_{u,i}^{\,q}\,\mathrm{LTV}_{u,i}\
  }{}{}
  \addConstraint{\sum_{u \in \mathcal{U}}\sum_{i \in \mathcal{I}} x_{u,i}
    \widetilde{\mathrm{pUnsub}}_{u,i}^{\,q}}{\le \tilde C}
  \addConstraint{\sum_{ i \in \mathcal{I}} x_{u,i}}{\le 1}{\forall\,u}
  \addConstraint{x_{u,i}}{\ge 0}{\forall\,u,i}
\end{maxi!}

The solution of this LP is an allocation decision $\mathcal{D}(q, \tau) = \{ x_{u,i}^{\mathcal{D}(q, \tau)} \}$ that is a function of both the model quality $q$ and the degree of exploration $\tau$. The allocation decision $\mathcal{D}(q, 0)$ is the solution we obtain under no exploration and serves as a baseline. As we vary the two parameters, we explore the behavior of $\mathcal{D}(q, \tau)$ and key downstream quantities:

$$ \mathrm{R}(q, \tau) = \sum_{u \in \mathcal{U}} \sum_{i \in \mathcal{I}} \textrm{pConv}_{u,i} \,\textrm{LTV}_{u,i } \,x_{u,i}^{\mathcal{D}(q, \tau)},$$

$$ \mathrm{U}(q, \tau) = \sum_{u \in \mathcal{U}} \sum_{i \in \mathcal{I}} \textrm{pUnsub}_{u,i}\,x_{u,i}^{\mathcal{D}(q, \tau)}.$$

These are total expected revenue and total expected unsubscriptions, respectively. Across multiple simulation runs, we also explore stability quantities such as:

$$\textrm{AV}(q, \tau) = \frac{1}{|\mathcal{U}| |\mathcal{I}|}\sum_{u \in \mathcal{U}} \sum_{i \in \mathcal{I}} \textrm{Var}\left(x_{u,i}^{\mathcal{D}(q,\tau)}\right) $$

This is the total average allocation variance of $\mathcal{D}(q, \tau)$ across multiple runs.

\end{document}